\documentclass[conference, a4paper]{IEEEtran}
\IEEEoverridecommandlockouts
\usepackage{cite}
\usepackage{amsmath,amssymb,amsfonts}
\usepackage{algorithmic}
\usepackage{graphicx}
\usepackage{textcomp}
\usepackage{xcolor}

\usepackage{bm}
\usepackage{booktabs}
\usepackage{multirow}
\usepackage{url}
\usepackage[hidelinks]{hyperref}

\def\BibTeX{{\rm B\kern-.05em{\sc i\kern-.025em b}\kern-.08em
    T\kern-.1667em\lower.7ex\hbox{E}\kern-.125emX}}
\begin{document}

\title{Efficiency without Compromise: CLIP-aided Text-to-Image GANs with Increased Diversity}

\author{\IEEEauthorblockN{1\textsuperscript{st} Yuya Kobayashi}
\IEEEauthorblockA{\textit{SonyAI} \\
u.kobayashi@sony.com}
\and
\IEEEauthorblockN{2\textsuperscript{nd} Yuhta Takida}
\IEEEauthorblockA{\textit{SonyAI} \\
yuta.takida@sony.com}
\and
\IEEEauthorblockN{3\textsuperscript{rd} Takashi Shibuya}
\IEEEauthorblockA{\textit{SonyAI} \\
takashi.tak.shibuya@sony.com}
\and
\IEEEauthorblockN{4\textsuperscript{th} Yuki Mitsufuji}
\IEEEauthorblockA{\textit{SonyAI, Sony Group Corp.} \\
yuhki.mitsufuji@sony.com}
}

\maketitle

\begin{abstract}
Recently, Generative Adversarial Networks (GANs) have been successfully scaled to billion-scale large text-to-image datasets. However, training such models entails a high training cost, limiting some applications and research usage. To reduce the cost, one promising direction is the incorporation of pre-trained models. 
The existing method of utilizing pre-trained models for a generator significantly reduced the training cost compared with the other large-scale GANs, but we found the model loses the diversity of generation for a given prompt by a large margin.
To build an efficient and high-fidelity text-to-image GAN without compromise, we propose to use two specialized discriminators with Slicing Adversarial Networks (SANs) adapted for text-to-image tasks. Our proposed model, called SCAD, shows a notable enhancement in diversity for a given prompt with better sample fidelity. We also propose to use a metric called Per-Prompt Diversity (PPD) to evaluate the diversity of text-to-image models quantitatively. 
SCAD achieved a zero-shot FID competitive with the latest large-scale GANs at two orders of magnitude less training cost.
\end{abstract}


\begin{figure*}[t]
    \centering
    \includegraphics[scale=1.0]{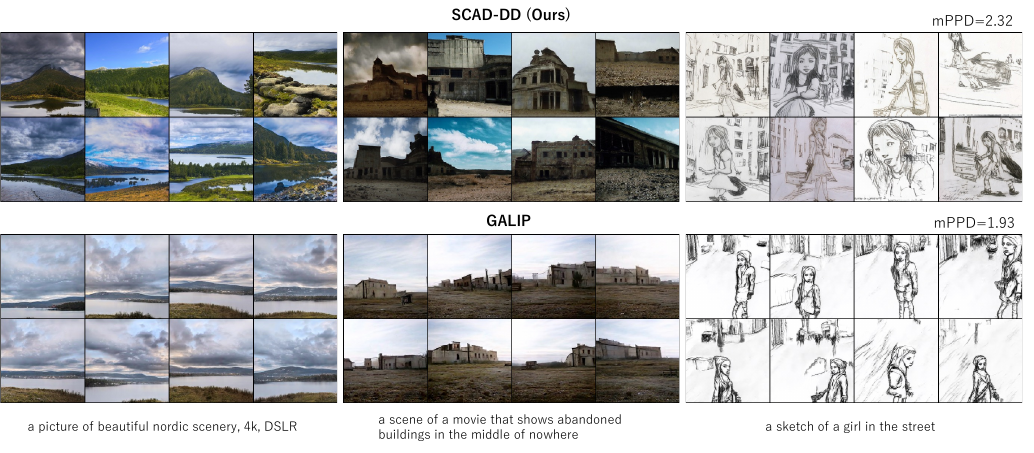}
    \caption{Images generated from GALIP and the proposed model SCAD-DD trained with CC12M dataset (in 224$\times$224). Samples from SCAD-DD have better image quality and diversity in terms of contents and styles. Samples of GALIP were obtained from the official checkpoint.}
    \label{fig:diversity_224}
\end{figure*}

\section{Introduction}
\label{sec:intro}

Despite the dominance of diffusion models\cite{ho2020ddpm, LDM2022, ramesh2022dalle2}, Generative Adversarial Networks (GANs)\cite{goodfellow2014gan} remain competitive in terms of sampling speed and fidelity of samples \cite{xiao2022DDGAN}. Several researchers successfully scaled GANs to about 1~billion parameters with $10^7$--$10^9$ text-image pairs and showed that they achieved a performance comparable to that of the latest diffusion models \cite{Sauer2023ICML, kang2023gigagan}. Thanks to the nature of the GANs' generator, which serves as a single neural function, their inference speed is up to hundreds of times faster than that of diffusion and autoregressive models\cite{kang2023gigagan}. Although GANs can sample images very fast, large-scale GANs come at a high training cost as well as the other types of large generative models do\cite{kim2023bksdm}. Expensive training can be a hindrance to both practical and research uses.

A promising way to reduce the training costs is to utilize self-supervised pre-trained models such as DINO\cite{caron2021emerging} and CLIP\cite{radford2021learning}. Incorporating pre-trained models into discriminators, which is called Projected GANs\cite{Sauer2021NEURIPS}, is a well-established practice. However, their use for generators remains less commonplace while they usually occupy a large portion of parameters in GANs. Tao et al.\ \cite{tao2023galip} adeptly integrated pre-trained CLIP into their model GALIP, achieving decent fidelity close to StyleGAN-T\cite{Sauer2023ICML} while using one order of magnitude less data (Tab.~\ref{tab:comparison256}). We call this type of generator as a CLIP-aided generator. Although this type of generator increases the training efficiency a lot, we find that it significantly reduces the diversity of generation for a given prompt. We call this problem conditional mode collapse. Insufficient diversity of generation can lead to averaged unattractive samples, and it also makes it difficult for us to choose satisfactory samples.

In this paper, we propose three methods to address conditional mode collapse and develop two variants of efficient text-to-image GANs by combining these methods. We observe that those methods are effective for conditional mode collapse while simultaneously improving the fidelity metrics. The first one is to use two expert discriminators. One is designed for text-alignment, and the other is for image fidelity. With some preliminary experiments, we noticed there is a trade-off between the diversity of generated images and text alignment or image quality, and introducing expert discriminators can alleviate this trade-off.

The second one is to adapt Slicing Adversarial Networks (SANs)\cite{takida2024san}. Takida et al.\ \cite{takida2024san} introduced three conditions that enable discriminators to properly estimate the distance between real and fake distributions through theoretical analysis, a property they named metrizability. Inspired by this study, we hypothesized one of the reasons for conditional mode collapse might be a rough estimation of the distance between distributions. Therefore, we extend SAN for text-to-image generation tasks, which has not yet been explored.

The third one is a regularization based on mutual information (MI)\cite{shannon1948, Kreer1957mutual} between input noise and a generated image. Input noise is the only source of randomness in GANs, so enhancing the MI between them can enrich diversity~\cite{belghazi2018mine}. As another motivation, we also expect MI regularizer to enhance metrizability according to the theoretical insight from~\cite{takida2024san}. We empirically show that MI regularization has synergy with the SAN-based loss and it enhances the fidelity metrics significantly.

Combining a CLIP-aided generator, SAN, MI regularization, and expert discriminators, we propose efficient GAN models that retains diversity for a prompt. We borrow some of the letters from these methods and term our base model as SCAD, and its two variants SCAD-MI and SCAD-DD. SCAD-DD achieves a zero-shot FID$_\text{30k}$ of 11.65 on COCO in about 190 A100 days. Also, SCAD-MI achieved a zero-shot FID$_\text{30k}$ of 13.93 in about $10$ A100 days, which is a comparable zero-shot FID$_\text{30k}$ to that of StyleGAN-T\cite{Sauer2023ICML} at 1\% of its training cost.

In our work, we also pose a question on the evaluation of the diversity of text-to-image generative models. While standard metrics such as Fréchet Inception Distance (FID)\cite{martin2017ttur} and Inception Score (IS)\cite{tim2016inception} consider diversity in unconditional generation tasks, we find that they are not effective for conditional generation tasks, especially for text-to-image generation tasks, since they do not evaluate the diversity of images from a single prompt. Hence, we define a metric, Per-Prompt Diversity (PPD), in Section~\ref{sec:ppd}.

We introduce the background and motivations of each method in Section~\ref{sec:motivations}, and the details of our methods and implementations in Section~\ref{sec:method}. Our contributions can be summarized as follows:
\begin{itemize}
    \item To alleviate the trade-off between the diversity of generated images and text-alignment or image quality in GANs, we introduce expert discriminators designed for different tasks.
    \item To address conditional mode collapse in the CLIP-aided generator, we extend Slicing Adversarial Networks (SANs) for text-to-image generation tasks.
    \item Our models (SCAD-MI and SCAD-DD) achieve richer diversity for a given prompt both qualitatively and quantitatively (e.g., in Fig.~\ref{fig:diversity_224} and Tab.~\ref{tab:results224}). For quantitative evaluation, we define a metric named Per-Prompt Diversity (PPD).
    \item SCAD-DD trained with the CC12M dataset achieves a zero-shot FID$_\text{30k}$ of 12.34 on the MS COCO. Also, SCAD-MI achieves 13.93, which is comparable to StyleGAN-T with less than 1\% of its training cost.
\end{itemize}

\section{Background}
\label{sec:motivations}
\label{sec:mtv_clip-aided}

Tao et al. achieved efficient training in GALIP by incorporating a pretrained CLIP module into the generator\cite{tao2023galip}. However, we observed that GALIP offers little diversity of generation for a given prompt. As mentioned in Section~\ref{sec:intro}, we call this type of generator a CLIP-aided generator, and denote the lack of diversity for a prompt as conditional mode collapse. We review these two topics in the following subsections. Conditional mode collapse has not been prominent in DF-GAN~\cite{tao2022df}, the predecessor of GALIP, which shares most of the architectures with GALIP except the CLIP-aided generator. Conditional mode collapse is also not present in other large-scale GANs such as StyleGAN-T\cite{Sauer2023ICML} or GigaGAN\cite{kang2023gigagan}. Therefore, we hypothesized that the CLIP-aided generator is the core of the problem, and there is room for improvement.

\subsection{CLIP-aided Generator}
\begin{figure}[t]
    \centering
    \includegraphics[scale=0.68]{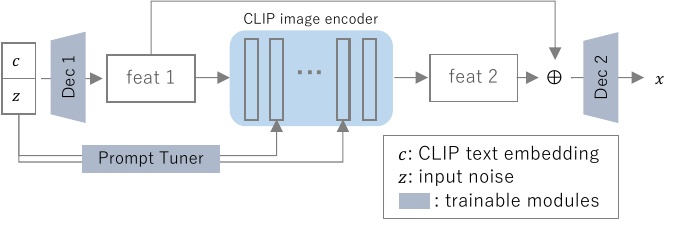}
    \caption{Overview of CLIP-aided generators. Two decoders (Dec1, Dec2) and the Prompt Tuner are trainable. CLIP image encoder (blue shaded part) is frozen.}
    \label{fig:generator}
\end{figure}

First, we review the architecture of the CLIP-aided generator. An overview of the generator is depicted in Fig.~\ref{fig:generator}. It is composed of a frozen CLIP image encoder and three trainable modules: two decoders and the Prompt Tuner. The first decoder (Dec1) was originally named a ``bridge feature predictor"~\cite{tao2023galip}, but we call it a ``decoder" in accordance with our interpretation, which is introduced later in this section. The first decoder (Dec1) converts the concatenation of an input noise $\bm z\in\mathbb{R}^{D_z}$ and a sentence embedding $\bm c\in\mathbb{R}^{D_{c}}$ to an image feature for the CLIP: feat 1. Then, another image feature (feat 2) is obtained via a pretrained CLIP image encoder. Note that the input convolution layer and the last projection layer of CLIP are omitted here. Finally, the main decoder (Dec2) generates an image $\bm{x}\in \mathbb{R}^{D_x}$ from these features. Here, the Prompt Tuner converts $\bm z$ and $\bm c$ into intermediate features, and the features are then concatenated into the intermediate features of the CLIP image encoder. This can be regarded as noise injection with text conditioning~\cite{stylegan, feng2021noiseinjection}.

Next, we share our view on how the CLIP-aided generator works. We interpret the CLIP-aided generator as a two-stage process. We consider the first image feature as a pseudo ``image prompt'' that extracts features from the CLIP encoder corresponding to the input text $\bm c$. 
CLIP's image embeddings are supposed to be similar regardless of the image quality or subtle difference in the pixel domain as long as the contents of the images are similar.
Therefore, the first feature can be coarse as long as it can be recognized by CLIP. This makes the task of the first decoder easy enough for stable and fast convergence, which is as if drafting. Then, the second feature can be regarded as a ``purified'' version of the first feature, which is more suitable for generation.

\subsection{Causes of Conditional Mode Collapse} \label{sec:collapse}
However, this two-stage process has a drawback in terms of the diversity of generation. As the second feature is well aligned with text condition $\bm c$---in other words, ``purified'' as we mentioned above---the feature tends to lose randomness. 
Thus, although the feat2 is supposed to be a good conditioning for image generation, if the second decoder depends on the feat2 too much, the resulting samples will be less diverse. This lack of the diversity is actually observed in GALIP, which can be seen in Fig.~\ref{fig:diversity_224} and Fig.~\ref{fig:ppd_coco}.

In addition, CLIP guidance on the generator is considered to be another cause of the conditional mode collapse. CLIP guidance is a widely used regularization in many GAN-based text-to-image models that enhances the CLIP score between generated images $\bm{x}$ and conditions $\bm c$. Thus, the emphasis on $\bm c$ relatively decreases the importance of $\bm z$, which is a source of randomness, for the decoder.

In this paper, we inherit the architecture of the CLIP-aided generator that is considered to play an important role in efficient training, and we introduce 1) Slicing Adversarial Networks (SANs)~\cite{takida2024san} loss adapted for text-to-image generation, 2) Expert discriminators to alleviate the trade-off between per-prompt diversity and sample quality, and 3) a mutual information regularizer. The details of each method are introduced in the following section.


\section{Methods}
\label{sec:method}

\subsection{Slicing Adversarial Network for Text-to-Image Synthesis}
Takida et al. \cite{takida2024san} proposed a modified training scheme for GANs based on sliced Wasserstein distance, which is named Slicing Adversarial Networks (SANs). SANs are shown to mitigate the issue that the discriminators in existing GANs fail to evaluate a distance between real and fake distributions~\cite{takida2024san}. They termed this property of discriminators that estimates the distance between distributions as ``metrizability'' and proposed three conditions to achieve it: injectivity, separability, and direction optimality. The objective function of SANs was proposed to achieve direction optimality among these conditions.

A possible cause of conditional mode collapse is the rough estimation of the distance between real and fake distributions by the discriminators. Inaccurate estimation of distance in high-dimensional space leads to some modes being ignored. While SAN loss was developed for unconditional generation and class-conditional generation in the original paper, it has not yet been explored for text-to-image generation. In this paper, we aim to extend SANs to the text-to-image generation tasks.

\subsubsection{SAN Loss}
We first review the SAN formulation for unconditional generation. In general, discriminators can be represented as
\begin{equation}
    f(\bm{x}) = \bm{w} \cdot \bm{h}(\bm{x}),
\end{equation}
where $\bm{x}\in\mathbb{R}^{D_{x}}$ is an input image, $\bm{w}\in\mathbb{R}^{D_{h}}$ represents a last linear projection
, and $\bm{h}:\mathbb{R}^{D_{x}}\to\mathbb{R}^{D_{h}}$ is a feature extractor that projects an input into $D_{h}$ dimensional features. The dot represents the Euclidean inner product. 

The following maximization objective is used in SANs to ensure the direction optimality in discriminators
\begin{align} \label{eq:uncondsan1}
        \mathcal{V}_\text{hinge}(\bm{h};\bm{\omega}) =& \mathbb{E}_{x\sim p_{\text{real}}}\left[ \min(0, -1+\bar{\bm{\omega}} \cdot \bm{h}(\bm{x}) \right] \\ \nonumber
        &+ \mathbb{E}_{x\sim p_{\text{fake}}}\left[ \min(0, -1-\bar{\bm{\omega}} \cdot \bm{h}(\bm{x})) \right], \\
        \mathcal{V}_\text{wass}(\bm{\omega};\bm{h}) =& \mathbb{E}_{x\sim p_{\text{real}}}\left[ \bm{\omega} \cdot \bar{\bm{h}}(\bm{x})\right] - \mathbb{E}_{x\sim p_{\text{fake}}}\left[ \bm{\omega} \cdot \bar{\bm{h}}(\bm{x}) \right], \label{eq:uncondsan2}
\end{align}
where $p_\text{real}$ is an empirical distribution given by a dataset, $p_\text{fake}$ is a distribution of generated images, and $\bar{(\cdot)}$ is a stop gradient operator. $\bm{\omega}$ is a normalized direction on a hypersphere, i.e., $\bm{\omega} \in \mathbb{S}^{D-1}$. This can be regarded as applying the hinge loss for $\bm{h}(\bm{x})$, and the Wasserstein loss for the direction $\bm{\omega}$. The overall maximization problem is a summation of them, namely, $\mathcal{V}_{\text{SAN}} = \mathcal{V}_{\text{hinge}} + \mathcal{V}_{\text{wass}}$.

SAN for class-conditional generation is also proposed by using as many directions $\bm{\omega}$ as the number of classes. Since text embeddings have continuous values, the same approach can not be applied to text-to-image generation.

\subsubsection{The Extension for Text-to-Image}
\label{sec:san_method}
\begin{table}[t]
    \centering
    \caption{Ablation study of discriminators with CUB dataset\cite{WahCUB_200_2011}. $\text{sn}[\cdot]$ and SN stand for spectral normalization. FIDs and CLIP scores are evaluated by GALIP's official implementation. $\bm c$ is a condition, and $\bm x$ is an image.}
    \begin{tabular}{@{}l|@{}cc}
         Architecture & FID & CLIP score \\
         \midrule
         GALIP(reproduced) & \ 10.08 & 0.3164 \\
         GALIP+SN & \ 10.10 & 0.3125 \\
         \midrule
         $\bm{h}(\bm{x}) \cdot \bm{\omega}$ & \ 11.47 & 0.3066 \\
         $\bm{h}(\bm{x}, \bm{c}) \cdot \bm{\omega}$ & \ 11.41 & 0.3070 \\
         $\bm{h}(\bm{x}, \bm{c}) \cdot \bm{\omega}(\bm{c})$ & \ 10.08 & 0.3100 \\
         $\text{sn}[\bm{h}](\bm{x}, \bm{c}) \cdot \text{sn}[\bm{\omega}](\bm{c})$ & \ 10.31 & 0.3127 \\
         $\text{sn}[\bm{h}](\bm{x}, \bm{c}) \cdot \bm{\omega}(\bm{c})$ & \ \bf{9.669} & \bf{0.3174}
    \end{tabular}
    \label{tab:disc_arch}
\end{table}
\begin{figure*}[t]
    \centering
    \includegraphics[scale=0.8]{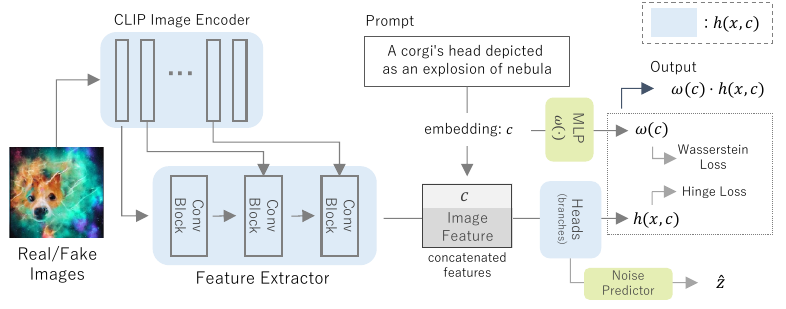}
    \caption{Overview of our discriminator. It incorporates an objective function originating from SAN. This illustration is based on the semantic branch mainly. The condition $c$ is not used in the case of fidelity branch. Noise Predictor is only for SCAD-MI.}
    \label{fig:overview}
\end{figure*}
There are several candidate formulations for extending SANs to text-to-image generation tasks. As for the direction $\bm{\omega}$, we expect the text conditional $\bm{\omega}(\bm{c})$ version to be a natural extension of the class-conditional case, but it can also be unconditional. The conditioning of $\bm{h}(¥\bm{x})$ has the same options as well. We conducted ablation studies on the CUB-200-2011 dataset\cite{WahCUB_200_2011} to find an optimal choice. The options and the results are enumerated in Tab.~\ref{tab:disc_arch}. We found the performance of conditioning both $\bm{h}$ and $\bm{\omega}$ by text $\bm c$ was the best.
Here, conditional version $\bm{\omega}(\bm{c})$ can be interpreted as preparing an infinite number of discriminators smoothed by $\bm{c}$, which is more efficient than preparing many discriminators individually.
As a result, the objective function extended for text-to-image generation can be written as
\begin{align} \label{eq:condsan}
    \mathcal{V}_\text{hinge}^\text{c}(\bm{h};\bm{\omega}) &= \mathbb{E}_{(\bm{x},\bm{c})\sim p_{\text{real}}}\left[ \min(0, -1+\bar{\bm{\omega}}(\bm{c}) \cdot \bm{h}(\bm{x},\bm{c}) \right] \\ \nonumber
    & + \mathbb{E}_{(\bm{x},\bm{c})\sim p_{\text{fake}}}\left[ \min(0, -1-\bar{\bm{\omega}}(\bm{c}) \cdot \bm{h}(\bm{x},\bm{c})) \right], \\
    \mathcal{V}_\text{wass}^\text{c}(\bm{\omega};\bm{h}) &= \mathbb{E}_{(\bm{x},\bm{c})\sim p_{\text{real}}}\left[ \bm{\omega}(\bm{c}) \cdot \bar{\bm{h}}(\bm{x},\bm{c})\right] \\  \nonumber
    & - \mathbb{E}_{(\bm{x},\bm{c})\sim p_{\text{fake}}}\left[ \bm{\omega}(\bm{c}) \cdot \bar{\bm{h}}(\bm{x},\bm{c}) \right].
\end{align}
The overview of our discriminator is depicted in Fig.~\ref{fig:overview}.

Moreover, in Tab.~\ref{tab:disc_arch}, we tested spectral normalization (SN) on the discriminator so that the map $\bm{h}$ does not collapse. As we expected, applying SN on discriminator base $\bm{h}$ enhanced the performance, but its application on the direction $\bm{\omega}$ slightly deteriorated the performance. 
We conjecture that an optimal direction $\bm{\omega}$ given $\bm c$ should be determined uniquely, but regularization like SN can interfere with it. As a result, we chose $\text{sn}[\bm{h}](\bm{x}, \bm{c})\cdot \bm{\omega}(\bm{c})$ as our model, where $\text{sn}[\cdot]$ is a spectral normalization operator. For comparison, we also applied SN to GALIP (second row in the table), but no performance gain was observed. 

In addition to this, a mismatch loss is inherited from GALIP \cite{wei2018improving, tao2022df}. For mismatch loss, augmented fake distribution is defined as $\tilde{p}_\text{fake} = (p_\text{fake}+p_\text{mis})/2$, where mismatch distribution $p_{\text{mis}}$ is an incorrect combination of text and image pairs. $\mathcal{V}_\text{hinge}^\text{c}(\bm{h};\bm{\omega})$ and $\mathcal{V}_\text{wass}^\text{c}(\bm{h};\bm{\omega})$ replaced with a $\tilde{p}_\text{fake}$ instead of a $p_{\text{fake}}$ are denoted as $\tilde{\mathcal{V}}_\text{hinge}^\text{c}(\bm{h};\bm{\omega})$ and $\tilde{\mathcal{V}}_\text{wass}^\text{c}(\bm{h};\bm{\omega})$ respectively.

\subsection{Expert Discriminators}
CLIP guidance is an essential technique to enhance text-image alignment in GANs; however, enhancing CLIP guidance can lead to less diversity as explained in Section~\ref{sec:collapse}. This trade-off can be controlled by some techniques such as a mutual information regularization we introduce in the next section. Although such kind of techniques are useful, they tend to end up with moderate improvement with slightly degrading other metrics.

To avoid the trade-offs, we need to help increasing text-alignment with a method other than the CLIP guidance. For this purpose, we introduce two expert discriminators (discriminator heads) specialized for image fidelity and text-alignment respectively. We call the former discriminator as a "fidelity branch", and the latter one as a "semantic branch". Discriminators in text-to-image GANs need to care about image fidelity and semantic alignment concurrently, which makes the discrimination task more complicated. Dividing these two tasks softly and preparing dedicated discriminators for each task are expected to make it easier and hence improve the text-alignment.


We adopted the architecture and the losses introduced in Section~\ref{sec:san_method} for the semantic branch. The objective is $\tilde{\mathcal{V}}_\text{hinge}^\text{c}(\bm{h};\bm{\omega})$ and $\tilde{\mathcal{V}}_\text{wass}^\text{c}(\bm{h};\bm{\omega})$. They take an image and a condition as an input, and mismatch loss are applied. On the other hand, the fidelity branch incorporates PatchGAN architecture \cite{CycleGAN2017, Esser_2021_CVPR} without using any information about a text condition. PatchGAN architecture has a limited receptive field and is expected to focus on local features. The objective of the fidelity branch is $\mathcal{V}_{\text{hinge}}$ and $\mathcal{V}_{\text{wass}}$ in eq.~\eqref{eq:uncondsan1} and \eqref{eq:uncondsan2}. Combined with the fidelity branch, semantic branch is expected to focus on discriminating text-image alignment mainly.

\subsection{Regularization by Mutual Information}
As explained in Section~\ref{sec:mtv_clip-aided}, we hypothesize that conditional mode collapse happens when the decoder ignores noise $\bm z$, which is similar to the posterior collapse in Variational Autoencoders~\cite{kingma2014vae, bowman2016posterior}. Another possible solution is to apply regularization of mutual information (MI) between noise and generated images $I(\bm x; \bm z)$. We regard the SAN-based loss and the expert discriminators introduced above as a remedy for conditional mode collapse from a discriminator's perspective, and MI regularization as another one mainly from a generator's perspective.

MI regularization has been utilized in some GAN-based models such as InfoGAN~\cite{xi2016infogan} and MINE~\cite{belghazi2018mine}. 
A novel mutual information estimator showing effectiveness for the prevention of mode collapse with small toy datasets was reported in \cite{belghazi2018mine}. Although our motivation is similar to this, we also use it for enhancing the metrizability, and we work on a large-scale real dataset.

We introduce the formulation and feasible approximation of the MI regularizer. Here, we denote a distribution of generated images by a generator $G$ as 
$\bm{x}, \bm{z} \sim p_\text{fake}(\bm{x}, \bm{z}) := \mathbb{E}_{\mathbf{c}\sim p_{c}}\left[ p_{\text{fake}}(\bm{x}|\bm{z}, \bm{c}) p(\bm{z})\right]$
, where $\bm{c}\sim p_c$ is the distribution of sentence embeddings from the CLIP text encoder, $p_\text{fake}(\bm{x}|\bm{z},\bm{c})$ is a generator distribution conditioned by $\bm{z}$ and $\bm{c}$, and $p(\bm{z})=\mathcal{N}(\bm{0}, \bm{I})$ is the prior of input noise $\bm{z}\in \mathbb{R}^{D_z}$. By its definition, MI can be decomposed as 
\begin{equation}
    I(\bm x; \bm z) = H(\bm z) - H(\bm z | \bm x), \label{eq:mi2}
\end{equation}
where $H$ indicates entropy. We used the variational approximation as in InfoGAN~\cite{barber2003mi}. The objective comes from $H(\bm{z}|\bm{x})$ in \eqref{eq:mi2}, as the entropy term $H(\bm{z})$ is constant in this case. It can be written as

\begin{equation}
    \mathcal{L}_{\text{MI}}=\mathbb{E}_{\bm{z}\sim p(\bm{z})}\left[{\mathbb{E}_{\bm{x}\sim p_{\text{fake}}(\bm{x} | \bm{z})}[\log Q(\bm{z}|\bm{x})]}\right],
\end{equation}
where $Q(\bm{z}|\bm{x})$ is an approximation function. This approximation function can be implemented by a noise prediction head with MSE loss attached to a discriminator.

The task of this estimator can be regarded as a reconstruction of $\bm{z}$ from sampled images $\bm{x}$, and this can prevent the collapse of maps $p_\text{fake}(\bm{x} | \bm{z})$ in the generator and $\bm{h}(\bm{x}, \bm{c})$ in the discriminator as well. Hence, we hypothesize that this also enhances injectivity on discriminators. Note that injectivity is one of the conditions for metrizable discriminators, as introduced in the previous section~\ref{sec:san_method}.

\subsection{Objective Functions} \label{sec:objective}

We also kept matching aware gradient penalty from GALIP. Combining all the losses introduced above, the resulting objective functions are
\begin{align}
    \underset{D}{\max}\ V(D;G) &= (\tilde{\mathcal{V}}_{\text{hinge}}^\text{c} + \tilde{\mathcal{V}}_{\text{wass}}^\text{c} - \mathcal{L}_\text{GP}^\text{c}) \nonumber \\
    &+ (\mathcal{V}_{\text{hinge}} + \mathcal{V}_{\text{wass}} - \mathcal{L}_\text{GP}) + \lambda\mathcal{L}_\text{MI}, \label{eq:max} \\
    \underset{G}{\min}\ V(G;D) &= \mathbb{E}_{\bm{z}\sim p_{z}, \bm{c}\sim p_{c}}\left.[D_{\text{sem}}(\text{CL}(G(\bm{z}, \bm{c})), \bm{c}) \right. \nonumber \\
    &+ D_{\text{fid}}(\text{CL}(G(\bm{z}))) \left. - \mu \cdot \text{CS}(G(\bm{z}, \bm{c}), \bm{c}) \right], \label{eq:min} 
\end{align}
where $\mathcal{L}_\text{GP}^\text{c}$ is a matching-aware gradient penalty, and $\mathcal{L}_\text{GP}$ is a standard gradient penalty:
\begin{align}
        \mathcal{L}_\text{GP}^\text{c} =& \mathbb{E}_{(\bm{x},\bm{c})\sim p_{\text{real}}}[k_1\left\| \nabla_{\text{CL}(\bm{x})} D(\text{CL}(\bm{x}), \bm{c})\right\|_{2}^{l_1} \nonumber \\
        & \qquad + k_2\left\| \nabla_{\bm{c}}[D(\text{CL}(\bm{x}), \bm{c})] \right\|_{2}^{l_2}], \\
        \mathcal{L}_\text{GP} =& \mathbb{E}_{\bm{x}\sim p_{\text{real}}}[k_1\left\| \nabla_{\text{CL}(\bm{x})} D_{\text{fid}}(\text{CL}(\bm{x}))\right\|_{2}^{l_1}].
\end{align}
Here, $D_{\text{sem}}(\text{CL}(\bm{x}),\bm{c})=\bm{\omega}(\bm{c})\cdot \bm{h}(\bm{x}, \bm{c})$ is a semantic branch, and $D_{\text{fid}}(\text{CL}(\bm{x})) = \bm{\omega} \cdot \bm{h}(\bm x)$ is a fidelity branch. $\text{CL}(\bm x)$ is a CLIP image encoder in $\bm{h}(\bm{x}, \bm{c})$, $G(\bm{z}, \bm{c})=p_\text{fake}$ is a generator, and $\text{CS}(G(\bm{z}, \bm{c}), \bm{c})$ in eq.~\eqref{eq:min} is a CLIP score for the CLIP guidance. 

The default values are $k_{1}=0.5$, $k_{2}=0.1$, $l_{1}=6$, $l_{2}=1$, and $\mu=4.0$. In this paper, we define a setting using single discriminator (only semantic branch) without mutual information loss ($\lambda=0.0$) as a base model SCAD. Incorporating the techniques introduced above, we define two variants named SCAD-MI (Mutual Information) and SCAD-DD (Dual Discriminators). SCAD-MI has a single discriminator (only semantic branch) and mutual information loss ($\lambda=1.0$). SCAD-DD has two expert discriminators without mutual information loss ($\lambda=0.0$). We do not introduce a method that combines both mutual information regularization and expert discriminators due to two reasons. The first one is that SCAD-DD achieved decent diversity and fidelity even without the mutual information regularization, and the other one is that we found combining them makes it difficult to tame training. Thus we decided to introduce a light-weight SCAD-MI and a better-performing SCAD-DD.

\section{Diversity Metric for Text-to-Image Generation}
\label{sec:ppd}
\subsection{Evaluation of Text-to-Image Models}
Widely used metrics such as Fréchet Inception Distance (FID)\cite{martin2017ttur} and Inception Score (IS)\cite{tim2016inception} are designed to evaluate the fidelity and diversity of generated images. IS evaluates diversity as the entropy of the predicted labels, and FID evaluates it by comparing variances of generated images and real images in the embedding space of Inception-V3. However, we pose a problem that existing metrics can not properly evaluate the diversity of text-to-image models. 

Although text-to-image models are expected to generate various images from a given prompt, only a single image is sampled from each prompt in the standard evaluation process. Arguably, diversity for a given prompt can not be evaluated in this way. Since collecting multiple images corresponding to a prompt on a large scale is virtually impossible, evaluating the diversity of text-to-image models by comparing real and fake images, as is the case with FID, is also challenging.
As for IS, using uniformity of pre-defined classes is impractical for the evaluation of the diversity of images sampled from the same prompt, as they should all be in the same class.

Therefore, we introduce an evaluation metric regarding diversity for a given prompt in text-to-image generation; Per-Prompt Diversity (PPD). This metric represents the dispersion of image embeddings from a pretrained model. A similar diversity metric using LPIPS \cite{zhang2018perceptual} has been proposed in the context of subject-driven generation \cite{ruiz2023dreambooth}. Their method requires quadratic time $\mathcal{O}(n^2)$ because it requires to compute LPIPS value for all combinations of the images. In contrast, PPD is designed for a relatively large number of images and requires only linear time $\mathcal{O}(n)$.

\subsection{Definition of Per-Prompt Diversity (PPD)}
\begin{figure}
    \centering
    \includegraphics[width=\linewidth]{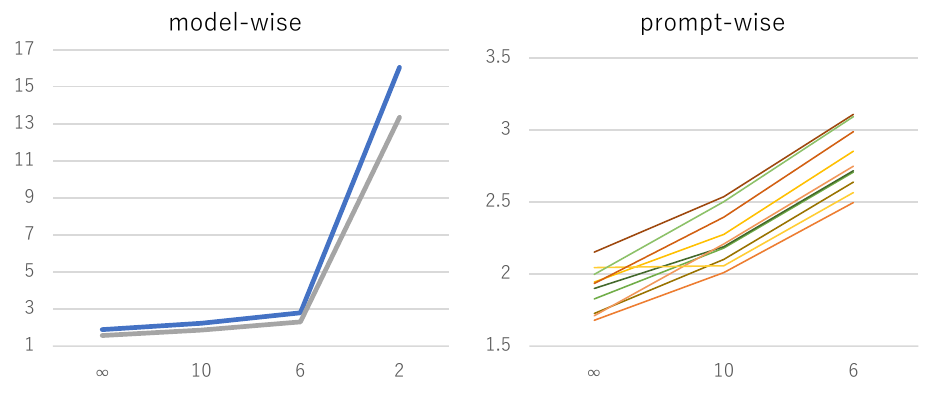}
    \caption{Validation of PPD with various $p$ (the order of a norm). The left figure shows the averaged PPD of 10 prompts in SCAD-MI (blue line) and GALIP (gray line). The right figure shows PPDs with different 10 prompts in SCAD-MI.}
    \label{fig:valid_ppd}
\end{figure}
We define Per-Prompt Diversity (PPD) using DINOv2 with register as an image encoder~\cite{oquab2024dinov2, darcet2024register}, denoted as $\bm{s}:\mathbb{R}^{D_x}\to \mathbb{R}^{768}$:

\begin{equation}
    \text{PPD}(\bm{c}) = \sum_{i=1}^N\|\bm{s}(G(\bm{z}_i,\bm{c}))-\bar{\bm{s}}_{\bm{c}}\|_p,
\end{equation}
where $\bar{\bm{s}}_{\bm{c}}$ is an average of $N$ embeddings. We tested multiple values of $p$ including infinite norm (Fig.~\ref{fig:valid_ppd}), and confirmed that the order is kept among different $p$ values, which means it is not a sensitive hyperparameter. We chose $p=10$ in our experiments in order not to underestimate distances in a high-dimensional embedding space. Higher PPD indicates better diversity.

For model-wise evaluation, we also propose the use of mean PPD (mPPD), which is an average of PPD for different $K$ prompts, $\bm{c}_1, \ldots, \bm{c}_K$.
This is similar to the CLIP score used for the evaluation of both a single data point and a model. We computed $N=40$ and $K=1000$ for GANs, and $N=20$ and $K=500$ for diffusion models to save computation. Unlike FID, using a larger number of evaluation samples does not lead to a better score; it only affects the accuracy of the estimation.

\begin{figure*}[t]
    \centering
    \includegraphics[width=\linewidth]{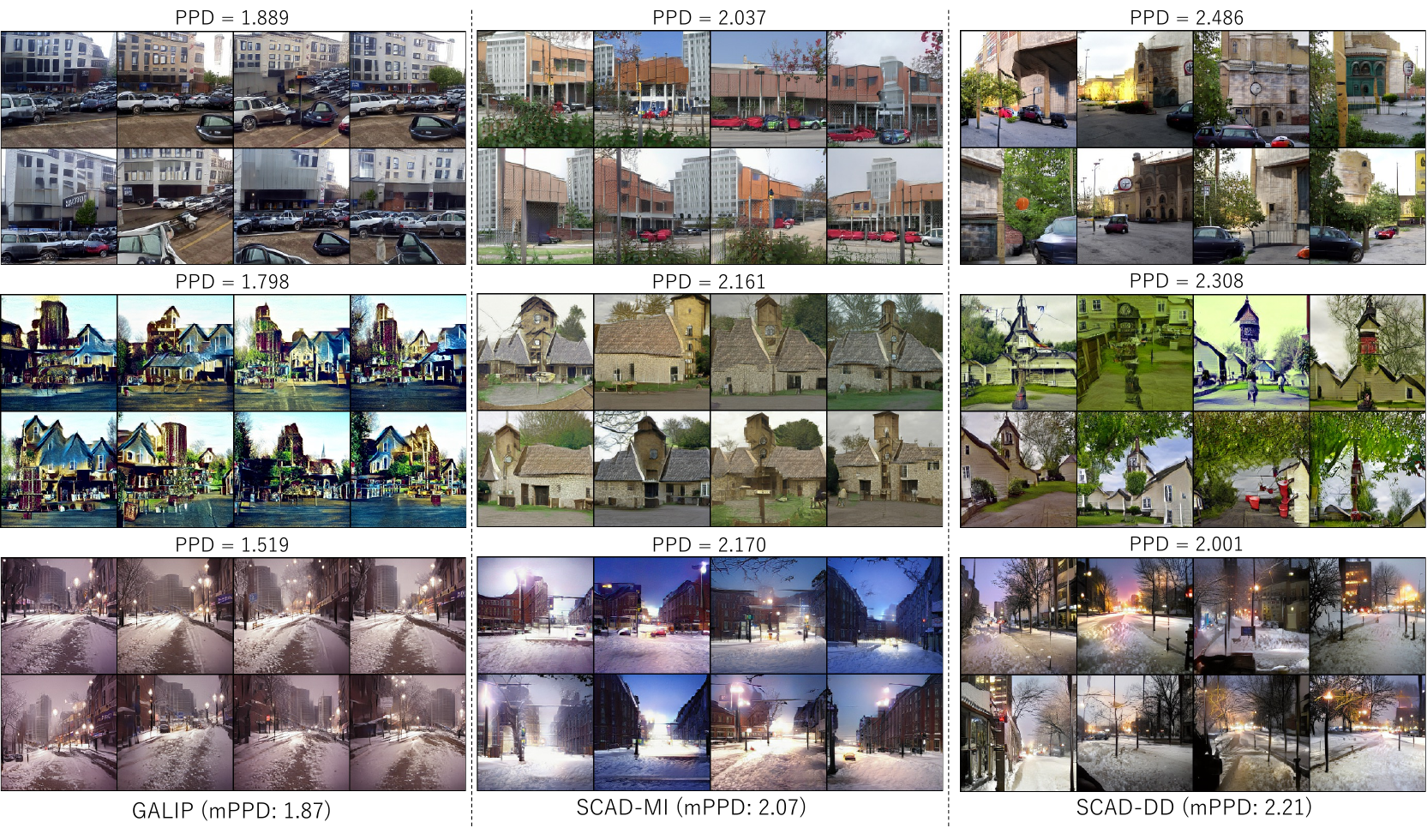}
    \caption{Samples from the models trained with the COCO dataset (83k images) in 224$\times$224. PPD of each prompt is shown above the images. Higher PPD indicates richer diversity. The prompts are from the validation set of the COCO dataset. Prompt1: ``a few cars that are parked next to a building'' Prompt2: ``The Oast House, The Street, Wittersham, Kent'' Prompt3: ``A city street covered in a light film of snow.''}
    \label{fig:ppd_coco}
\end{figure*}

\begin{table*}[t]
    \centering
    \caption{Model comparison in $224\times224$ resolution. $\text{FID}^\text{CLIP}$ is a variant of a FID using CLIP\cite{kynkaanniemi2023the}. zFID$_\text{30k}$ stands for zero-shot FID on the COCO dataset using 30k images for the evaluation. The error on mPPD indicates a standard error.}
    \resizebox{2\columnwidth}{!}{
    \begin{tabular}{l||cccc|ccccc}
        \toprule
        \multirow{2}{*}{Model} & \multicolumn{4}{c}{COCO} & \multicolumn{4}{c}{CC12M}  \\
         & FID$_\downarrow$ & ${\text{FID}^\text{CLIP}}_\downarrow$ & CLIP Score $_\uparrow$ & mPPD $_\uparrow$ & zFID${_{\text{30k}}}_\downarrow$ & zFID${_{\text{30k}}^{\text{CLIP}}}_\downarrow$ & CLIP Score & mPPD $_\uparrow$ & A100 days$_\downarrow$\\
        \midrule
        GALIP & 5.938 & 2.66 & \textbf{0.3587} & 1.87\scriptsize{$\pm0.0072$} & 13.33 & 7.104 & \textbf{0.3355} & 1.93\scriptsize{$\pm 0.0067$} & 30 \\
        SCAD (ablation) & 5.647 & 2.050 & 0.3533 & 2.15\scriptsize{$\pm0.0067$} & -- & -- & -- & -- & -- \\    
        SCAD-MI (ours) & \textbf{5.243} & 1.874 & 0.3471 & 2.07\scriptsize{$\pm{0.0068}$} & \textbf{10.90} & 5.896 & 0.3304 & 2.28\scriptsize{$\pm 0.0061$} & 30 \\
        SCAD-DD (ours) & 5.433 & \textbf{1.642} & 0.3511 & \textbf{2.21}\scriptsize{$\pm{0.0067}$} & 11.25 & \textbf{5.423} & 0.3284 & \textbf{2.32} \scriptsize{$\pm 0.0058$} & 60\\
        \bottomrule
        
    \end{tabular}
    }
    \label{tab:results224}
\end{table*}

\section{Experiments}
\label{sec:results}
\subsection{Settings}
We implemented our model based on GALIP's official implementation\footnote{\url{https://github.com/tobran/GALIP}} in PyTorch. We used NVIDIA V100, A100 and H100 GPUs, and we basically used BFLOAT16 (BF16) for training our models except for some ablation studies. BF16 generally makes the training numerically unstable, but can reduce memory consumption.

\subsubsection{Evaluation}
Since original GALIP generates images in $224\times224$ resolution, we first trained our model with this resolution. We also conducted experiments using typical $256\times256$ resolution for fair comparison with other generative models.

We found that GALIP's evaluation script tends to output lower (better) FID compared to the common libraries such as clean-fid\footnote{\url{https://github.com/GaParmar/clean-fid}} and torch-fidelity\footnote{\url{https://github.com/toshas/torch-fidelity}}. We also confirmed that the results from these two libraries were very close. In this paper, we used 30k images for zero-shot FID (zFID$_\text{30k}$), and all the 41k images for models trained with the COCO dataset directly. We used clean-fid in all the evaluations. For CLIP scores, we used the original CLIP from OpenAI (ViT-B/32)\footnote{\url{https://github.com/openai/CLIP}} basically, but in Tab.~\ref{tab:comparison256} we used OpenCLIP (ViT-G/14)\cite{openclip} to compare with other GANs.

\subsubsection{Normalization}
\label{sec:norm}
GALIP restricted the input features fed into CLIP to positive values, but this is not the original design of CLIP. We found that using the standard normalization enhances the diversity, but it worsens FID, and makes the training slightly unstable. We used the default normalization in our model in favor of naturalness of the method, and because FID and training stability was no longer a problem for our models. Our interpretation is that restricting the input to positive values makes the input features less informative for CLIP. Therefore it worked as a kind of truncation technique, which basically enhances image quality and degrades diversity.

\subsection{Ablation: COCO Dataset}
\label{sec:result_coco}
We trained our models with the COCO dataset\cite{mscoco} in $224\times224$ resolution. Training duration is about 1500~epochs with a batch size of $512$. We show the contribution of each module with the COCO dataset first. Then the zero-shot evaluation with a large-scale dataset (CC12M\cite{changpinyo2021cc12m}) is shown in the next section. 

\subsubsection{Qualitative Evaluation}
Firstly, we show samples from each model in Fig.~\ref{fig:ppd_coco}. The values above the images are the PPD introduced in Section~\ref{sec:ppd}, which represents the diversity of the images from the prompt. Each set of eight images was sampled from the same set of randomly sampled noises $\bm{z}$ for a fair comparison. We can see limited diversity with the images from GALIP. The samples from SCAD-MI and SCAD-DD are significantly more diverse in accordance with the higher PPD. They have various layouts and lightings. Note that these models are trained with only 83k images (COCO dataset) for an ablation study, and not for the best image quality.

\subsubsection{Quantitative Evaluation}
To assess the contribution of each method qualitatively, we evaluated the models with the metrics shown in Tab.~\ref{tab:results224}. All the trainings and evaluations were done in 224$\times$224 resolution. FID$^{\text{CLIP}}$ is an updated version of FID using CLIP, which is reported to be a more robust metric\cite{kynkaanniemi2023the}. 

Our models achieved more than 10\%-20\% improvement on mPPD from GALIP, which is perceptually noticeable as shown in Fig.~\ref{fig:ppd_coco}. As for the CLIP score, the trade-offs between the CLIP score and diversity or between the CLIP score and fidelity has been observed in text-conditional generation~\cite{ho2021cfg, Sauer2023ICML}. Although the CLIP score of GALIP is slightly better than the others, it is partly due to the limited diversity (low mPPD).

\begin{table*}[t]
    \centering
    \caption{Quantitative evaluation of the models on COCO in $256\times256$ resolution. SDv1.4: Stable Diffusion v1.4. $^*$:~Evaluated in $256$px by downsampling. \, $^{**}$:~A union of multiple datasets. \, $^{***}$:~Scores are from the official checkpoint or the paper. FID was evaluated in 256px by upsampling. mPPD was evaluated in native resolution: 224px. \, $^\dag$:~24 3090 days, which is almost identical to A100 days in TF32\, $^\ddag$:~Trained on BF16, H100 GPUs. A100 days are approximation. SCAD-MI in 256px has larger generator. The sign ``(early)'' indicates the results with shorter training.}
    \label{tab:comparison256}
    \begin{tabular}{l|ccccccc}
        \toprule
        Model & zFID${_{30k}}_\downarrow$ & zFID${_{30k}}^{\text{CLIP}}_\downarrow$ & mPPD$_\uparrow$ & CLIP Score$_\uparrow$ & A100 days & \#params & \#data \\
        \midrule
        SDv1.4 & 12.63 & - & 3.065\scriptsize{$\pm 0.0041$} & 0.308 & 6250 & 1.45B & 5B (LAION-5B)\\
        \midrule
        GigaGAN$^*$ & 9.09 & - & - & 0.307 & 4,783 & 1.0B & $>$2B (mixed$^{**}$)\\
        StyleGAN-T & 13.90 & - & - & 0.283 & 1,706 & 1.123B & 250M (mixed$^{**}$)\\
        GALIP$^{***}$ & 14.55 & 8.73 & 1.93\scriptsize{$\pm 0.0067$} & 0.293 & 25$^{\dag}$ & 0.32B & 12M (CC12M)\\
        SCAD-MI (early)$^\ddag$ & 13.93 & 8.576 & 2.26\scriptsize{$\pm 0.0062$} & 0.272 & 10 & 0.288B & 12M (CC12M)\\
        SCAD-DD (early)$^\ddag$ & 13.78 & 7.557 & 2.28\scriptsize{$\pm 0.0056$} & 0.271 & 100 & 0.201B & 12M (CC12M)\\
        \textbf{SCAD-MI (ours)}$^\ddag$ & 12.40 & 7.301 & 2.25\scriptsize{$\pm 0.0062$} & 0.281 & 30 & 0.288B & 12M (CC12M)\\
        \textbf{SCAD-DD (ours)}$^\ddag$ & 12.34 & 6.531 & 2.26\scriptsize{$\pm 0.0059$} & 0.276 & 190 & 0.201B & 12M (CC12M)\\
        \bottomrule
    \end{tabular}
    \label{tab:my_label}
\end{table*}

\begin{figure*}[t]
    \centering
    \includegraphics[width=\linewidth]{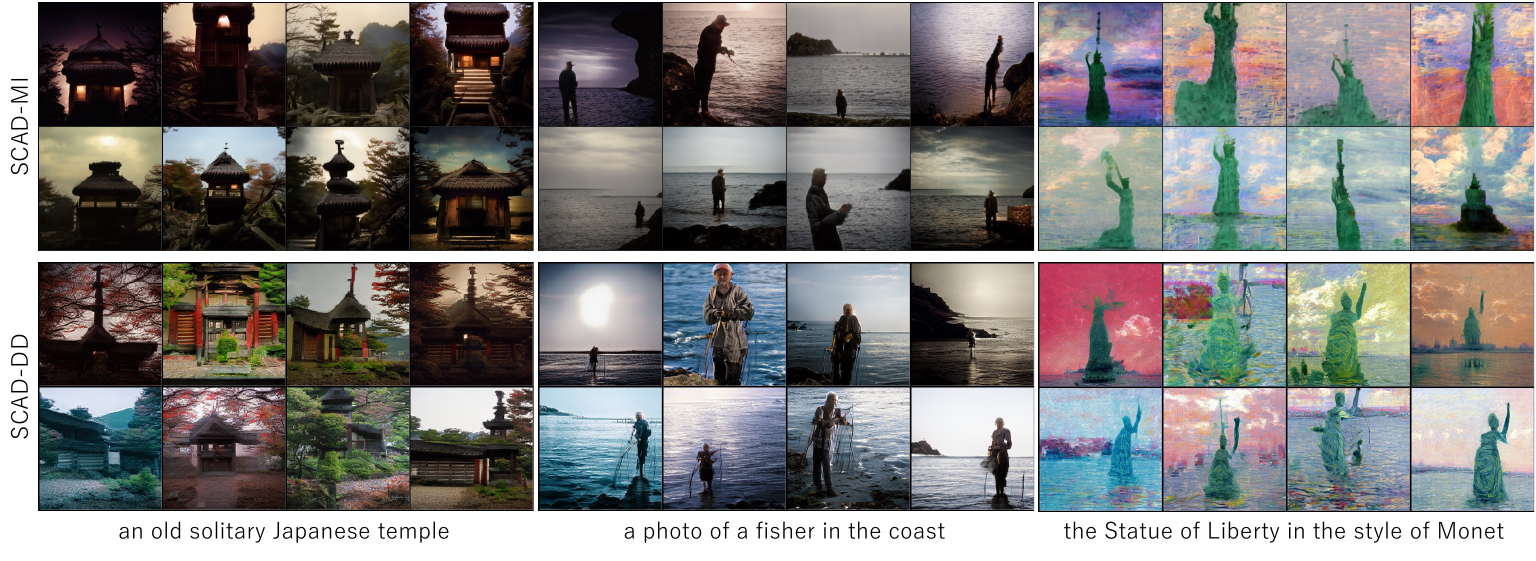}
    \caption{256$\times$256 images generated by our proposed models. An suffix ", movie light, high-resolution, best quality" is added to the left two prompts, and ", high-resolution, masterpiece, artstation" is added to the right prompt.}
    \label{fig:scad256}
\end{figure*}

\subsection{Zero-shot Performance: CC12M Dataset}
\subsubsection{Evaluation in $224\times224$}
In this section, we trained the proposed model in both $224\times224$ and $256\times256$ resolution with CC12M~\cite{changpinyo2021cc12m}, a more diverse, large-scale dataset. The $224$px resolution is for comparison with GALIP, and the $256$px is for other generative models since many other models are evaluated in this resolution.

The results in 224px are shown on the right side of Tab.~\ref{tab:results224}. SCAD-DD gained more than 20\% improvement both in zero-shot FID$^{\text{CLIP}}$ and in diversity (mPPD). As we can see from this table, mPPDs with CC12M are better than the ones with COCO. Using a larger dataset seems to make mPPD higher and CLIP score lower. 

Qualitative results can be seen in Fig.~\ref{fig:diversity_224}. The samples from GALIP have little diversity, and the content seems to be simpler compared to SCAD-DD. This is probably because the model is only able to generate averaged images according to a given condition (text prompt). The samples from SCAD have more variations in terms of components, styles, and layouts, which usually leads to better perceptual quality. 

\subsubsection{Evaluation in $256\times256$ (Full Model)}
Table.~\ref{tab:comparison256} shows the results in 256px. This table shows zero-shot FID$_{\text{30k}}$ (zFID$_\text{30k}$), zero-shot FID$_{\text{30k}}^{\text{CLIP}}$ (zFID$_\text{30k}^\text{CLIP}$), mPPD, CLIP Score, A100 days required for the training, the number of parameters, and the size of the training dataset. All the CLIP Scores in this table are calculated using OpenCLIP (ViT-G/14)\cite{openclip}. SCAD-DD achieved the zFID$_\text{30k}$ of 12.34 and the zFID$_\text{30k}^{\text{CLIP}}$ of 6.531 with 190 A100 days (the bottom row) with decent diversity (mPPD). Although the resulting FID itself is not the best compared to the latest diffusion models and the largest GAN (GigaGAN), SCAD deserves attention considering the amount of resources required for the training. 

In the early stage of the training, SCAD-MI (early) achieved the zFID$_\text{30k}$ of 13.93, which is almost same as StyleGAN-T in score-wise only with 10 A100 days and 12 million text-image pairs. This is computationally very efficient, and will be useful in projects without a big budget. We did not evaluate mPPDs of StyleGAN-T and GigaGAN because no implementations are available for GigaGAN, and the official StyleGAN-T implementation\footnote{\url{https://github.com/autonomousvision/stylegan-t}} is reported to be not reproducible\footnote{\url{https://github.com/autonomousvision/stylegan-t/issues/20}}. We also could not reproduce the results similar to the paper.

\begin{figure*}
    \centering
    \includegraphics[width=0.8\linewidth]{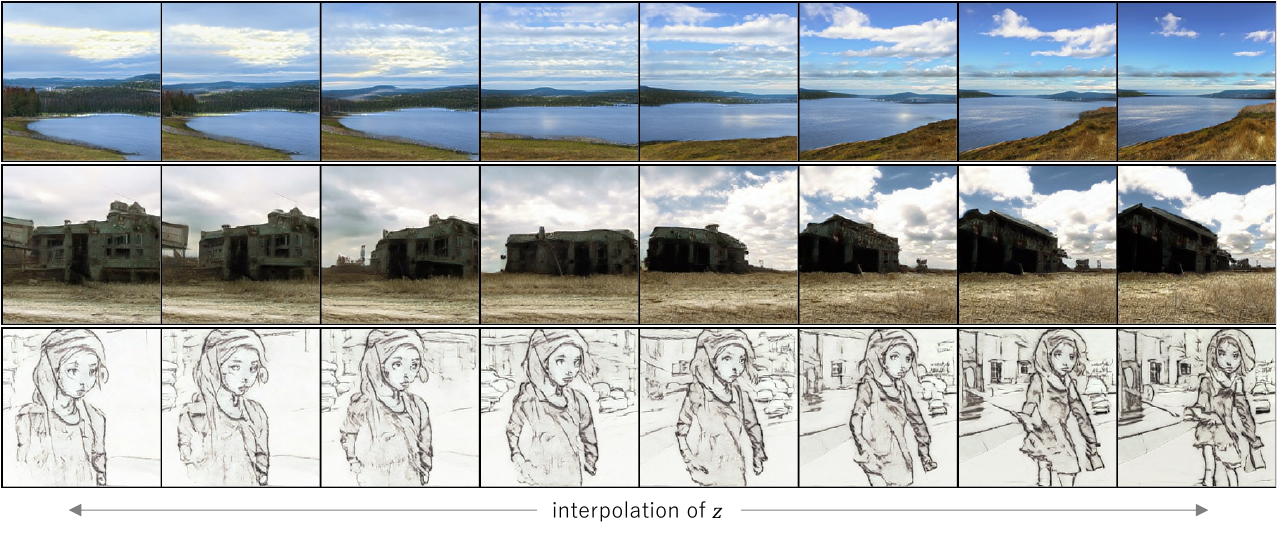}
    \caption{Interpolation of the latent space with SCAD-DD in $256\times256$. The same prompts as in Fig.~\ref{fig:diversity_224} are used.}
    \label{fig:interpolation}
\end{figure*}

Qualitative results from SCAD-MI and SCAD-DD trained with CC12M are shown in Fig.~\ref{fig:scad256} and in Fig.~\ref{fig:interpolation}. In Fig.~\ref{fig:scad256}, samples from SCAD-DD show slightly better details and variation than SCAD-MI. 
Fig.~\ref{fig:interpolation} shows linear interpolation of $\mathbf{z}$ in the latent space of SCAD-DD using the same prompts in Fig.~\ref{fig:diversity_224}. We can see that each set of images shows similar transition of a layout, which means the latent space is aligned with some attributes (e.g., layouts and color). This property can help reduce artifacts since it ensures a smooth transition in the pixel space.

\section{Related Work}
\label{sec:related_works}

\subsection{GANs for Text-to-Image Synthesis}
Text-to-image generation by GANs has been explored in various studies \cite{scott2016text-to-image, zhang2017stackgan, tao2018attngan, zhang2021xmcgan, zhou2022lafite, tao2022df}, but scaling GANs to large-scale text-image pairs has been challenging due to the training instability. Recently, many techniques to stabilize adversarial training have been explored such as gradient penalty\cite{wei2018improving}, spectral normalization\cite{miyato2018spectral}, progressive or hierarchical training\cite{zhang2017stackgan, karras2018progressive}, and projected discriminators\cite{Sauer2021NEURIPS}. By combining modern techniques, several studies have successfully scaled GANs to 1 billion parameters with hundreds of millions of text-image pairs\cite{Sauer2023ICML, kang2023gigagan}. Nevertheless, large-scale models incur very high training costs, making them prohibitive for iterative research processes. 

GALIP has achieved scaling to 12 million text-image pairs with relatively low training costs and no costly progressive training. However, as we mentioned in Section~\ref{sec:intro}, there is a significant problem of low diversity in the generated samples. This study aims to solve this issue.

\subsection{Evaluation Metrics}
The simplest metric for the evaluation of image generation quality is a log-likelihood of the model itself on a test dataset. However, some models e.g., GANs, does not compute likelihood explicitly.

The de facto standard metrics are sample-based metrics such as Fréchet Inception Distance (FID)\cite{martin2017ttur} and Inception Score (IS)\cite{tim2016inception}. 
FID and IS measure both image fidelity and diversity as a scalar. 
In addition, some other metrics have been developed for covering the areas in which they fall short, such as Kernel Inception Distance \cite{bińkowski2018demystifying}, precision/recall\cite{sajjadi2018precision, Kynkaanniemi2019precision}, Perceptual Path Length\cite{zhang2018perceptual} and Feature Likelihood Score \cite{jiralerspong2023feature}. 

However, these methods do not assume text-to-image generation and do not generate multiple images from a single prompt in the evaluation process. Hence, this approach can not address diversity for a given prompt inherently. Although FID can evaluate the diversity of generated images, it requires a ground truth image set. It is virtually impossible to collect multiple images for the same prompt on a large scale. Lee et al.\cite{lee2023holistic} discussed how generative models should be evaluated, but a variation of generation in text-to-image generation tasks is not included in it. In this study, we introduce a metric to evaluate the diversity of images for a given prompt quantitatively in Section~\ref{sec:ppd}.

\section{Conclusion}
In this paper, we proposed GAN-based models SCAD-MI and SCAD-DD. The models alleviated the conditional mode collapse observed in the CLIP-aided generator, and SCAD-DD achieved a zero-shot FID$_{30k}$ of 12.34 on the COCO dataset with only about 190 A100 days while keeping decent diversity. The components we proposed can also be beneficial for other GAN-based text-to-image models.

We also highlighted a problem in the evaluation of text-to-image synthesis, namely that the diversity of text-to-image models is difficult to measure with standard metrics such as FID and IS. To address this, we proposed per-prompt diversity (PPD) to evaluate the diversity of the generation for a given prompt quantitatively. In comparison with the latest generative models, the proposed models are small and trained with relatively small datasets. Evaluating the scalability of our method is a future direction.

\bibliographystyle{ieeetr}
\bibliography{egbib}

\end{document}